\documentclass[10pt,twocolumn,letterpaper]{article}
\usepackage{cvpr}
\usepackage{booktabs}
\usepackage{tabularx}
\usepackage{multirow}
\usepackage{amsmath}
\usepackage{amssymb}
\usepackage{graphicx}
\usepackage{microtype}
\usepackage{xspace}
\usepackage{siunitx}

\newcommand{\benchmark}{Safe2Drive\xspace}
\newcommand{\benchmarkshort}{S2D\xspace}

\newcommand{\ds}{DS\xspace}
\newcommand{\sr}{SR\xspace}

\definecolor{cvprblue}{rgb}{0.21,0.49,0.74}
\usepackage[pagebackref,breaklinks,colorlinks,allcolors=cvprblue]{hyperref}

\def\paperID{0000} 
\def\confName{CVPR}
\def\confYear{2026}

\title{Safe2Drive: Evaluating Safe Driving Behaviors of E2E Autonomous Driving Models}

\author{
Nishad Sahu$^{1}$ \quad Kalpana Panda$^{2}$ \quad Congyuan Yu$^{1}$\\
Changzhong Qian$^{1}$ \quad Shounak Sural$^{1}$ \quad Ragunathan Rajkumar$^{1}$\\
$^{1}$Carnegie Mellon University \quad $^{2}$Birla Institute of Technology and Science Pilani
}

\begin{document}
\maketitle

\begin{abstract}
Recent end-to-end (E2E) autonomous driving policies achieve high driving scores in closed-loop simulations. Yet it remains unclear whether these policies handle common safety-critical scenarios. We present \benchmark~(\benchmarkshort), a set of Bench2Drive-aligned scenario extensions focused on three frequent families of road hazards: work zones, pedestrian jaywalking, and occluded vulnerable road users (VRUs). \benchmark adds 100 common but challenging scenarios and introduces \emph{SafeDriving Score} (SDS), a safety-centric metric that augments prior evaluators with pre-crash braking, work zone-object contact, lane centering, and smoothness checks. Evaluating two state-of-the-art policies (LEAD and SimLingo) on \benchmarkshort, we find that their driving scores drop sharply relative to their reported Bench2Drive baselines (LEAD: from \SI{94.7}{\ds} on Bench2Drive to \SI{39.95}{\ds} on \benchmarkshort; SimLingo: from \SI{85.07}{\ds} on Bench2Drive to \SI{41.00}{\ds} on \benchmarkshort) and that SDS on \benchmarkshort is low (11.85 for LEAD and 15.27 for SimLingo). These results are consistent with brittle safe-driving behaviors such as poor work-zone understanding, red-light violations, and late or absent braking for pedestrians. This study highlights a lack of safe behavioral reasoning in E2E models even when tested on CARLA towns that are part of the training set. We plan to release the code and videos for all 100 \benchmarkshort scenarios.
\end{abstract}
\vspace{-6mm}
\section{Introduction}
Closed-loop evaluation remains the most credible way to assess end-to-end driving because the policy must recover from its own actions instead of merely replaying logged futures~\cite{jia2024bench2drive,nguyen2025lead}. Bench2Drive and CARLA Leaderboard 2.0, built on the CARLA simulator~\cite{dosovitskiy2017carla,carla_leaderboard}, made this setting substantially more reproducible by standardizing routes, infractions, and agent interfaces~\cite{jia2024bench2drive}. Yet strong aggregate scores still leave an important safety question unresolved: \emph{Does an E2E model understand safety-critical scene semantics and avoid violating safety constraints?}

\begin{figure}[t]
    \centering
    \includegraphics[width=\columnwidth]{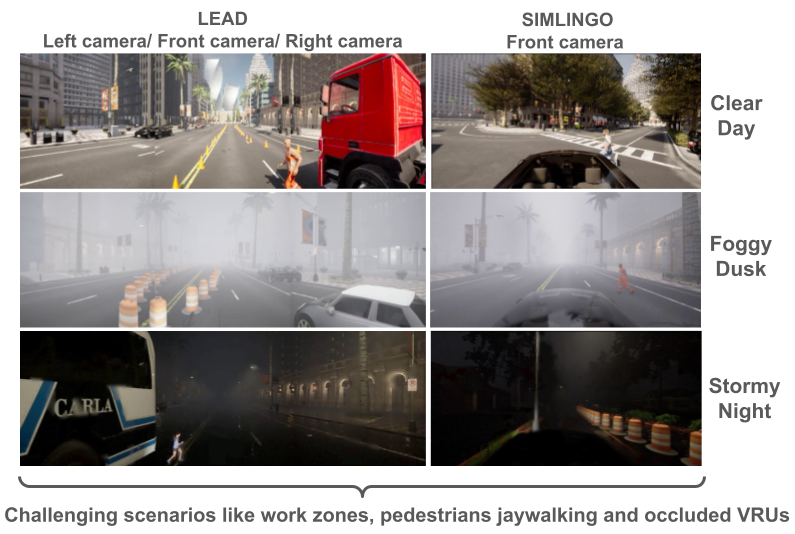}
    \caption{LEAD and SimLingo ego-view scenes while testing with S2D scenarios. The figure illustrates the three common hazard families emphasized in \benchmark: work zones, pedestrian jaywalking, and occluded VRUs. The rows show the same benchmark family under clear-day, foggy-dusk, and storm-night conditions.}
    \label{fig:c2d_teaser}
    \vspace{-6mm}
\end{figure}

We study this safe-driving question through three common scenario families: work zones, pedestrian jaywalking, and occluded vulnerable road users (VRUs). These are routine urban situations in which the policy must reason about lane reinterpretation, unsignaled pedestrian entry, and partial observability around vehicles and corners. Figure~\ref{fig:c2d_teaser} previews these scenario classes and shows that the same semantic hazards recur across both model interfaces and multiple weather regimes.

We introduce \benchmark~(\benchmarkshort), a Bench2Drive-style scenario extension for common behavioral safety cases. It currently covers work zones, pedestrian jaywalking, and occluded-VRU interactions at blind corners, T-intersections, and turning conflicts, while staying on the specified Bench2Drive evaluator path. We use the publicly available LEAD and SimLingo checkpoints~\cite{nguyen2025lead,renz2025simlingo} and test them on the S2D scenarios.

S2D consists of 100 scenarios: 75 work-zone routes in CARLA Town01--Town10, designed based on guidelines from a state transportation authority and comprising lane changes, merges, exits within work zones, detours, and forks; 10 jaywalking scenarios; and 15 occluded-VRU routes, including blind turning corners, a child occluded by a bus, and a construction worker occluded by a construction vehicle in a work zone. Quantitatively, the performance of both LEAD and SimLingo drops sharply from their reported Bench2Drive baselines: LEAD performance falls from \SI{94.7}{\ds}/\SI{82.1}{\sr} to \SI{39.95}{\ds}/\SI{37.00}{\sr} on \benchmark, and SimLingo falls from \SI{85.07}{\ds}/\SI{67.27}{\sr} to \SI{41.00}{\ds}/\SI{48.00}{\sr}. Under our proposed SafeDriving Score, SDS is 11.85 for LEAD and 15.27 for SimLingo on these 100 S2D scenarios. The primary insight, however, is not only the score gap but also clear behavioral safety concerns. Both models struggle to reinterpret narrowed work-zone corridors, respond unsafely to late pedestrian entry, and show significant red-light violations even in clear sunny weather.

This paper makes three major contributions.
\begin{itemize}
\item We introduce \benchmark~(\benchmarkshort), a Bench2Drive-style scenario extension for common behavioral safety evaluation, and create a 100-scenario set: 75 work-zone routes, 10 pedestrian-jaywalking routes, and 15 occluded-VRU routes. The work-zone family spans common transportation-authority-style layouts including lane changes, merges, detours, and exits within active work zones.
\item We analyze safety violations, including collisions and red-light infractions, to diagnose what the model could actually observe but still reason or act unsafely in jaywalking, occluded-VRU, and work-zone scenarios.
\item We propose SafeDriving Score (SDS), which explicitly adds penalties for work zone-object collisions, failing to decelerate before a collision, route-aware lane-corridor deviation, and multiplicative smoothness over short trajectory segments.
\end{itemize}

\section{Benchmark and Setup}
\textbf{Scenario family.} \benchmark is a scenario extension to Bench2Drive-style evaluation. We analyze a 100-route subset for which both LEAD and SimLingo artifacts exist in the repository: 75 work-zone routes, 10 pedestrian-jaywalking routes, and 15 occluded-VRU routes.

\textbf{Route sets.} Work-zone scenarios are defined by the presence of static work-zone objects (e.g., barrels), based on transportation-authority-inspired work-zone layouts. Jaywalking routes feature a child and an adult pedestrian running right in front of the ego vehicle to cross the road, as well as a pedestrian standing at an intersection. The occluded-VRU subset includes 15 routes including a running construction worker occluded by a construction vehicle, a child occluded by a stationary bus, blind-corner routes, and right-building blind spots. All routes include dynamic traffic, including emergency vehicles such as police cars and ambulances.

\textbf{Benchmark alignment.} All routes are executed through the Bench2Drive evaluator path. Work zones rely on a custom \texttt{StaticWorkzoneRoute} scenario with managed standard traffic. Jaywalking and occluded-VRUs use the same route and infraction machinery, with scenario-side actor trajectories for pedestrians and occluders. We however do \textit{not} change route-completion, infraction, or score-aggregation logic of these scenarios. 

\textbf{Traffic and weather.} Work zones include static work-zone objects along with managed background vehicles and pedestrians. The jaywalking and occluded-VRU subsets use scenario-specific pedestrians and, where relevant, explicit occluder vehicles or buildings. Weather is evaluated under \texttt{clear\_day}, \texttt{foggy\_dusk}, and \texttt{storm\_night} conditions where those archived runs exist.

\begin{table*}[t]
    \centering
    \scriptsize
    \setlength{\tabcolsep}{5pt}
    \resizebox{\textwidth}{!}{%
    \begin{tabular}{ll l cc cc ccc}
        \toprule
        \multirow{2}{*}{Model} & \multirow{2}{*}{Sensors} & \multirow{2}{*}{Backbone} & \multicolumn{2}{c}{Bench2Drive} & \multicolumn{2}{c}{S2D} & \multicolumn{3}{c}{S2D (Safe Driving)} \\
        & & & DS & SR & DS & SR & DS\textsubscript{safe}& Comfort & SDS \\
        \midrule
        LEAD & 3$\times$RGB + 2$\times$LiDAR + 4$\times$radar & TFv6 ResNet-34 & 94.70 & 82.10 & 39.95 & 37.00 & 36.88 & 0.321 & 11.85 \\
        SimLingo & vision-only RGB cameras & InternViT-300M-448px + Qwen2-0.5B & 85.07 & 67.27 & 41.00 & 48.00 & 37.58 & 0.406 & 15.27 \\
        \bottomrule
    \end{tabular}
    }
    \vspace{-3mm}
    \caption{Model-level comparison between official Bench2Drive and Safe2Drive 100-route \benchmark~(\benchmarkshort) subset. LEAD Bench2Drive numbers come from the reported TFv6 ResNet-34 checkpoint in~\cite{nguyen2025lead}. SimLingo Bench2Drive numbers come from the reported mean closed-loop Bench2Drive result in~\cite{renz2025simlingo}. \benchmarkshort~DS/SR are recomputed from the paired official-evaluator artifacts in this repository. SafeDriving columns report the mean route-level safe driving score, mean comfort score, and their product for S2D.}
    \label{tab:model_compare}
\end{table*}

\begin{table*}[t]
    \centering
    \tiny
    \setlength{\tabcolsep}{5pt}
    \resizebox{\textwidth}{!}{%
    \begin{tabular}{lcrrrrrr}
        \toprule
        Scenario type& Routes & LEAD DS & LEAD SR & LEAD Coll. & SimLingo DS & SimLingo SR & SimLingo Coll. \\
        \midrule
        Work zones & 75 & 39.11 & 34.67 & 36.00 & 33.29 & 40.00 & 45.33 \\
        Pedestrian jaywalking & 10 & 40.64& 40.00& 70& 55.76& 60& 50\\
        Occluded VRUs & 15 & 43.66& 46.67& 60& 69.71& 80& 40\\
        Combined& 100 & 39.95 & 37.00 & 43.00 & 41.00 & 48.00 & 45.00 \\
        \bottomrule
    \end{tabular}
    }
    \vspace{-3mm}
    \caption{Scenario-type-wise comparison on S2D. Collision rate is the percentage of routes with at least one official collision event.}
    \label{tab:benchmark}
    \vspace{-6mm}
\end{table*}

\textbf{Models.} We evaluate the public LEAD TFv6 ResNet-34 checkpoint with three RGB cameras, LiDAR, and radar, without retraining or fine-tuning on \benchmark~\cite{nguyen2025lead}. We compare it against archived evaluator SimLingo runs. SimLingo is a vision-only VLA model built on InternViT-300M-448px and Qwen2-0.5B~\cite{renz2025simlingo}. We restrict all multi-model comparisons to routes for which both models have checkpoint, metric, and input-video artifacts.

\textbf{SafeDriving Score (SDS).} Bench2Drive reports route completions and infractions based on a driving score, but it does not directly account for whether the ego vehicle attempted to brake before a collision, whether work zone control objects should be treated as safety-critical contacts, or whether a route was completed with unstable lane-centering and limited accountability for poor smoothness. We therefore define a safe driving score
\begin{equation}
\mathrm{DS}^{(r)}_{\text{safe}} =
\mathrm{DS}^{(r)}_{\text{route}}
\Bigl(\prod_{e \in \mathcal{I}_r} p(e)\Bigr)
0.8^{N^{(r)}_{\text{nodecel}}}
0.95^{N^{(r)}_{\text{anom}}}
0.9^{N^{(r)}_{\text{crit}}}
\vspace{-2mm}
\end{equation}

where $\mathcal{I}_r$ is the set of standard B2D infractions for route $r$, $N_{\text{nodecel}}$ counts collisions with no longitudinal deceleration in the preceding 20 frames, and $N_{\text{anom}}$/$N_{\text{crit}}$ count route-corridor segments whose lateral error excess exceeds \SI{0.25}{m} or \SI{0.65}{m}. We treat work zone-object collisions, including traffic cones and similar control devices, as a dedicated work zone infraction with penalty factor $0.5$, which is harsher than the standard layout-collision treatment because these contacts usually indicate unsafe corridor interpretations rather than incidental curb touches.

To handle driving comfort, we divide the trajectory into 20-frame segments and require six smoothness signals to remain within bounds: longitudinal acceleration, lateral acceleration, jerk magnitude, longitudinal jerk, yaw acceleration, and yaw rate. The route-level comfort score is given by 
\begin{equation}
C^{(r)} = \frac{1}{M_r} \sum_{m=1}^{M_r}
\mathbf{1}\!\left[
\bigwedge_{k=1}^{6}\; \max_{t \in \mathcal{S}_{r,m}} \bigl| s_k^{(r)}(t) \bigr| \le \tau_k
\right],
\end{equation}
where $M_r$ is the number of segments for route $r$, $\mathcal{S}_{r,m}$ is the set of timesteps in segment $m$ of route $r$, $s_k^{(r)}(t)$ denotes the $k$-th smoothness signal at timestep $t$ (longitudinal acceleration, lateral acceleration, jerk magnitude, longitudinal jerk, yaw acceleration, and yaw rate), and $\tau_k$ is its corresponding per-signal bound. At the benchmark level, we report
\begin{equation}
\mathrm{SDS} = \overline{\mathrm{DS}_{\text{safe}}} \times \overline{C},
\vspace{-3mm}
\end{equation}

that is, the mean safe driving score times the mean comfort score across the evaluation set. We compute SDS from the same archived checkpoint and telemetry artifacts used for the official results, with local OpenDRIVE maps for Town01 and Town10HD\_Opt stored along with our results for reproducibility.

\textbf{Interpretation and caveat.} SDS is intended as a complementary metric, not a replacement for the standard Bench2Drive DS. Its value is that SDS encodes safety judgments missing from the B2D driving score. We therefore report SDS as a safety-focused companion metric whose purpose is to reveal behavior that B2D DS can underweight, while retaining Bench2Drive DS/SR for primary benchmark comparability.


\begin{table*}[t]
    \centering
    \tiny
    \setlength{\tabcolsep}{5pt}
    \resizebox{0.75\textwidth}{!}{%
    \begin{tabular}{llcccc}
        \toprule
        Model & Weather & Red light & Layout collision & Outside-route lanes & Vehicle blocked \\
        \midrule
        LEAD & clear day & 3/34 (8.82\%) & 9/34 (26.47\%) & 8/34 (23.53\%) & 11/34 (32.35\%) \\
        LEAD & foggy dusk & 0/33 (0.00\%) & 10/33 (30.30\%) & 7/33 (21.21\%) & 13/33 (39.39\%) \\
        LEAD & storm night & 0/33 (0.00\%) & 7/33 (21.21\%) & 6/33 (18.18\%) & 9/33 (27.27\%) \\
        \midrule
        SimLingo & clear day & 4/34 (11.76\%) & 6/34 (17.65\%) & 6/34 (17.65\%) & 6/34 (17.65\%) \\
        SimLingo & foggy dusk & 0/33 (0.00\%) & 4/33 (12.12\%) & 7/33 (21.21\%) & 4/33 (12.12\%) \\
        SimLingo & storm night & 2/33 (6.06\%) & 16/33 (48.48\%) & 12/33 (36.36\%) & 9/33 (27.27\%) \\
        \bottomrule
    \end{tabular}
    }
    \vspace{-3mm}
    \caption{Weather-linked infraction patterns on the 100 S2D scenarios. Each entry is \emph{routes with the infraction / total routes} with route-level occurrence rate in parentheses. Daytime emphasizes traffic-light compliance failures; storm night emphasizes corridor-keeping and contact failures, especially for SimLingo; foggy dusk emphasizes blocking for LEAD.}
    \label{tab:weather_infractions}
    \vspace{-3mm}
\end{table*}

\section{Evaluation Results}
\begin{figure*}[t]
    \centering
    \includegraphics[width=0.98\textwidth]{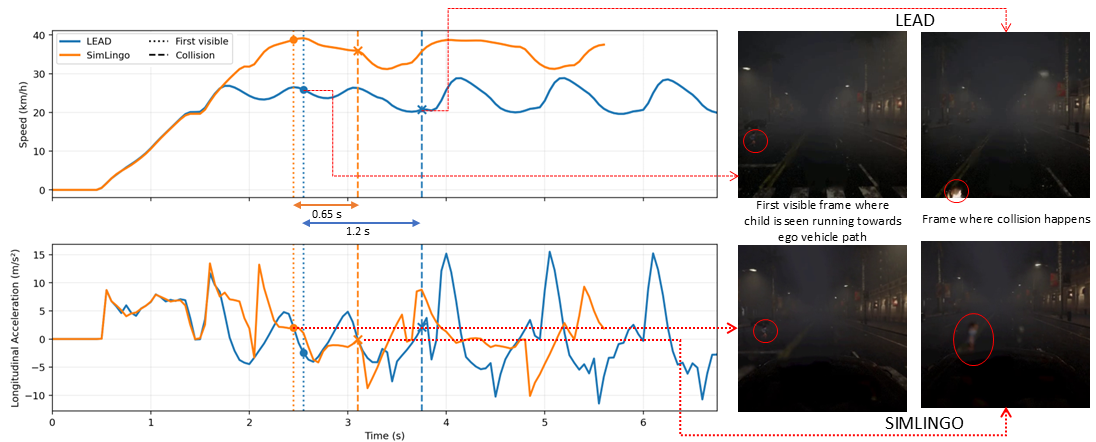}
  \vspace{-2mm}
    \caption{Collision analysis for a child jaywalking case in stormy night in S2D. The speed and longitudinal-acceleration traces are aligned with manually audited visibility and collision timestamps. LEAD first sees the child at \SI{2.55}{s} and collides at \SI{3.75}{s}; SimLingo first sees the child at \SI{2.45}{s} and collides at \SI{3.10}{s}. Even after the child becomes visible in the camera stream, neither model applies sustained braking through impact.}
    \label{fig:collision_analysis}
    \vspace{-6mm}
\end{figure*}

Table~\ref{tab:benchmark} shows the route-wise comparison of DS (Driving Score), SR (Success Rate), and Coll. (Collision Rate). On the 75 work-zone routes, both LEAD and SimLingo degrade sharply. LEAD averages \SI{39.11}{\ds} and \SI{34.67}{\sr}; SimLingo averages \SI{33.29}{\ds} and \SI{40.00}{\sr}. Table~\ref{tab:model_compare} places those \benchmarkshort\ numbers next to each model's reported Bench2Drive baseline. LEAD starts from the stronger official B2D baseline, but both models lose substantial performance once evaluation shifts toward common work-zone, jaywalking, and VRU-occlusion scenarios. The Safe Driving columns reinforce that conclusion. LEAD reaches a mean DS\textsubscript{safe} of 36.88 (without comfort multiplication), mean comfort of 0.321, and SDS of 11.85. SimLingo reaches a mean DS\textsubscript{safe} of 37.58, mean comfort of 0.406, and SDS of 15.27. Both SDS values are much lower than the official \benchmarkshort\ DS because many routes that still receive nonzero B2D driving score exhibit abrupt acceleration/deceleration, unstable lane keeping, or collisions without meaningful pre-impact braking.

Table~\ref{tab:weather_infractions} shows that different weather conditions tend to trigger additional kinds of safety violations rather than simply making all routes harder. Daytime exposes traffic-light compliance errors most clearly: LEAD records all three of its red-light-violation routes in \texttt{clear\_day}, and SimLingo records four of its six red-light-violation routes there as well. The absolute counts are small, but the concentration is notable because these failures happen under nominal daytime perception rather than only under degraded visibility, suggesting a semantic decision-making weakness in distinguishing when a visible signal still requires stopping.

Stormy nights, in contrast, expose corridor-keeping and contact failures most strongly for SimLingo. Its layout-collision rate rises from 17.65\% in \texttt{clear\_day} and 12.12\% in \texttt{foggy\_dusk} to 48.48\% in \texttt{storm\_night}, while outside-route-lane infractions rise to 36.36\%. This pattern is consistent with poor work-zone corridor interpretation and degraded lane discipline under adverse nighttime conditions. LEAD does not show the same stormy-night surge in layout or lane infractions, but it does show a different adverse-weather failure mode, mainly because LEAD uses radars and LiDARs whereas SimLingo uses cameras only. Thus, to ensure better performance under poor weather and visibility conditions, LiDAR and radar may be necessary in addition to cameras.

For LEAD, \texttt{foggy\_dusk} produces the highest blocked-route rate at 39.39\%, higher than \texttt{clear\_day} (32.35\%) and \texttt{storm\_night} (27.27\%). That suggests hesitation and stalled progress in moderate-visibility conditions rather than direct collision failures. Together, these trends reinforce the broader point of \benchmark: weather conditions matter, and they change \emph{which} behavioral weakness dominates, not just how often the model fails.


Figure~\ref{fig:collision_analysis} shows an S2D scenario on a stormy night in which a child is running across the road. The child is visible in the camera view before impact in both models, but neither responds with a clean, sustained deceleration to a safe stop. LEAD oscillates between braking and re-acceleration after first visibility and still hits the child at \SI{20.7}{km/h}. SimLingo maintains a much higher approach speed, sees the child \SI{0.65}{s} before impact, and still collides at \SI{35.9}{km/h}.


Taken together, these results show that high Bench2Drive scores do not correspond to safe driving outcomes on common safety-critical interactions.
\vspace{-4mm}
\section{Conclusion}
\vspace{-2mm}
We introduced \benchmark, a safety-centric scenario extension to Bench2Drive-style evaluation. We identified and analyzed a 100-route subset with paired LEAD and SimLingo artifacts: 75 work-zone routes, 10 pedestrian-jaywalking routes, and 15 occluded-VRU routes. Using the standard evaluator path, we found serious safety concerns in common work-zone, jaywalking, and occlusion-heavy scenes. We also introduced Safe Driving Score, which explicitly penalizes no-brake collisions, work-zone-object contact, route-corridor drift, and poor smoothness.

While the quantitative drop in driving score and driving safety is important, the nature of the failure cases is also noteworthy. Both LEAD and SimLingo struggle when work zones redefine lane geometry. Jaywalking scenarios show that the route-completion tally alone can hide unsafe late responses to pedestrian entry. Both models exhibit most of their red-light violations during daytime, indicating that these E2E models can struggle to account for red and green traffic signals under clear conditions compared with night and dusk. Moreover, SimLingo shows a major increase in collisions in storm-night conditions compared with LEAD, suggesting that camera-only E2E models degrade more under poor weather and visibility than models that also use radar and LiDAR.

Our next step is to extend S2D to other families of safety scenarios and address driving-safety concerns in E2E stacks related to emergency-vehicle interactions, school-bus stop-arm scenarios, more diverse blind-spot cases, and other challenging scenarios. This will ensure that safe driving behavior can be measured on a wider set of legally salient scene classes and compared across more than one end-to-end stack.

{\small
\bibliographystyle{ieeenat_fullname}
\bibliography{main}
}

\end{document}